\documentclass[10pt,twocolumn,letterpaper]{article}

\usepackage[pagenumbers]{cvpr}
\usepackage[accsupp]{axessibility} %
\usepackage{graphicx}
\usepackage{capt-of}
\usepackage{xcolor}
\usepackage{amsmath,amssymb,amsfonts,dsfont,pifont,bm,bbm,mathrsfs,mathtools,nicefrac}
\usepackage{algorithm,algpseudocode,listings}
\usepackage{booktabs,multirow,adjustbox,diagbox,threeparttable}
\definecolor{citeblue}{RGB}{48,111,186}
\usepackage[pagebackref,breaklinks,colorlinks,citecolor=citeblue,bookmarks=false]{hyperref}
\usepackage[capitalize]{cleveref}
\usepackage{url}
\crefname{section}{Sec.}{Secs.}
\Crefname{section}{Section}{Sections}
\crefname{table}{Tab.}{Tabs.}
\Crefname{table}{Table}{Tables}
\crefname{figure}{Fig.}{Figs.}
\Crefname{figure}{Figure}{Figures}
\crefname{equation}{Eq.}{Eqs.}
\Crefname{equation}{Equation}{Equations}
\usepackage[linewidth=1pt]{mdframed}
\usepackage{colortbl}
\definecolor{mygray}{gray}{.9}
\definecolor{mypink}{rgb}{.99,.91,.95}
\definecolor{mycyan}{cmyk}{.3,0,0,0}
\usepackage{array}
\definecolor{CQColor}{rgb}{0.0,0.0,1.0} %

\definecolor{CQRColor}{rgb}{1.0,0.0,1.0} %

\definecolor{CQXXYColor}{rgb}{1.0,0.0,0.0} %

\usepackage{bbding}
\usepackage{cite}
\definecolor{emphasizeColor}{HTML}{2755FF}

\definecolor{baseColor}{rgb}{0.75,0.05,0.1}

\definecolor{checkmarkColor}{rgb}{0.1,0.75,0.1}

\usepackage{enumitem}
\usepackage{makecell}
\usepackage{tabulary}
\definecolor{demphcolor}{RGB}{144,144,144}

\definecolor{mygray}{gray}{0.4}
\usepackage{pifont}
\newlength\savewidth\newcommand\shline{\noalign{\global\savewidth\arrayrulewidth
  \global\arrayrulewidth 1pt}\hline\noalign{\global\arrayrulewidth\savewidth}}

\newcommand{\tablestyle}[2]{\setlength{\tabcolsep}{#1}\renewcommand{\arraystretch}{#2}\centering\footnotesize}
\usepackage{xspace}

\usepackage{makecell}
\usepackage{tabulary}
\usepackage{enumitem}
\usepackage{arydshln}
\usepackage{ulem}
\definecolor{gdcolor}{RGB}{101.0,162.0,62.0} %

\usepackage{pifont}
\newcommand{\cmark}{\ding{51}}%
\newcommand{\fivestaropen}{\text{\ding{73}}}
\usepackage[pagebackref,breaklinks,colorlinks]{hyperref}
\usepackage[capitalize]{cleveref}
\crefname{section}{Sec.}{Secs.}
\Crefname{section}{Section}{Sections}
\Crefname{table}{Table}{Tables}
\crefname{table}{Tab.}{Tabs.}

\newcommand{\tocite}[1]{\textcolor{red}{[TO CITE]}}
\newcommand{\method}{VoP\xspace}

\begin{document}
\title{VoP: Text-Video Co-operative Prompt Tuning for Cross-Modal Retrieval}
\author{Siteng Huang$^{1, 3}$\thanks{Work done during internship at Alibaba DAMO Academy.}, Biao Gong$^{2}$, Yulin Pan$^{2}$, Jianwen Jiang$^{2}$, Yiliang Lv$^{2}$, Yuyuan Li$^{3}$, Donglin Wang$^{1}$\thanks{Corresponding author.}\\
{$^1$Machine Intelligence Lab (MiLAB), AI Division, School of Engineering, Westlake University}\\
{$^2$Alibaba Group}\ \ {$^3$Zhejiang University}\\
{\tt\small \{huangsiteng, wangdonglin\}@westlake.edu.cn, y2li@zju.edu.cn,}\\
{\tt\small a.biao.gong@gmail.com}, {\tt\small \{yanwen.pyl, jianwen.jjw, yiliang.lyl\}@alibaba-inc.com}
}

\maketitle

\begin{abstract}

Many recent studies leverage the pre-trained CLIP for text-video cross-modal retrieval by tuning the backbone with additional heavy modules, 
which not only brings huge computational burdens with much more parameters, 
but also leads to the knowledge forgetting from upstream models.
In this work, we propose the \method: Text-Video Co-operative Prompt Tuning
for efficient tuning
on the text-video retrieval task.
The proposed \method is an end-to-end framework with both video \& text prompts introducing, which can be regarded as a powerful baseline with only 0.1\% trainable parameters.
Further, based on the spatio-temporal characteristics of videos, we develop three novel video prompt mechanisms to improve the performance with different scales of trainable parameters.
The basic idea of the \method enhancement is to model the frame position, frame context, and layer function with specific trainable prompts, respectively.
Extensive experiments show that compared to full fine-tuning, the enhanced \method achieves a 1.4\% average R@1 gain across five text-video retrieval benchmarks with 6$\times$ less parameter overhead.
The code will be available at \url{https://github.com/bighuang624/VoP}.

\end{abstract}

\section{Introduction}

Due to the remarkable progress in large-scale contrastive language-image pre-training~\cite{Radford:CLIP,Li:DeCLIP,Jia:ALIGN,Li:ALBEF}, a recent popular direction for the crucial text-video cross-modal retrieval~\cite{Gabeur:MMT,Liu:multi-experts-VTR,Wang:object-aware-pretraining,Wu:HANet} task is to transfer pre-trained image-text knowledge to the video domain~\cite{Luo:CLIP4Clip,Zhao:CenterCLIP,Gorti:X-Pool} with fine-tuning.
However, the dominant full fine-tuning strategy inevitably forgets the useful knowledge acquired in the large-scale pre-training phase and poses a risk of overfitting, 
as the entire model is updated with limited downstream data.
Moreover, full fine-tuning requires to maintain an independent model weight for every dataset during deployment, which becomes infeasible due to the increasing model capacity.

\begin{figure}[!t]
  \centering
   \includegraphics[width=\linewidth]{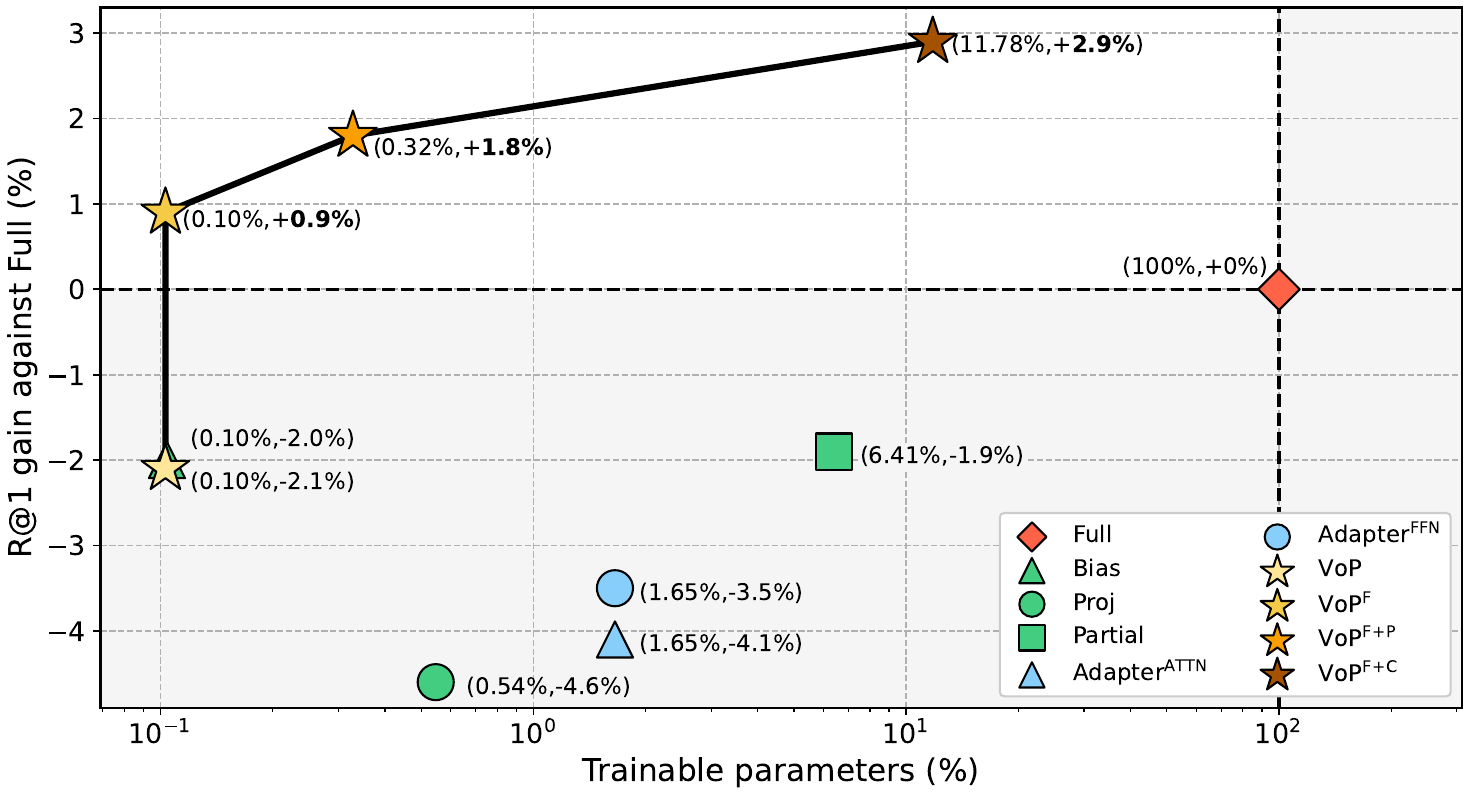}
   \vspace{-6mm}
   \caption{
   \textbf{Fine-tuning comparison of our proposed methods and full fine-tuning (Full)}.
   For each method, we represent the R@1 (recall at rank 1) gain on the MSR-VTT-9k dataset together with the number of trainable parameters.
   And we show only a part of our proposed methods for clarity, labeled with $\fivestaropen$.
   More detailed results are reported in \cref{sec:main_results}.}
   \label{fig:intro}
   \vspace{-2mm}
\end{figure}

In this paper, we introduce \textit{prompt tuning}~\cite{Liu:prompt-tuning,Lester:prompt-tuning} to address the challenges that limit the transferability and generalizability.
Keeping the backbone frozen and only tuning a few extra parameters prepended to the input, prompt tuning has been widely applied as a flexible and light-weight fine-tuning protocol.
Compared to uni-modal applications~\cite{Liu:P-Tuning-deep,Bahng:visual-prompting}, text-video cross-modal retrieval requires more parameters to support the dual-branch structure, making it logical to benefit from the parameter-efficient tuning strategy.
In addition, different from text descriptions that compose sequential information from words, video-understanding requires summarizing information in both the spatial and temporal dimensions.
Therefore, we assume that designing non-trivial video prompts further contributes to prompting both branches for mutual promotion.

According to the above discussion,
we propose the \textbf{\method}: Text-\textbf{V}ideo C\textbf{o}-operative \textbf{P}rompt Tuning to simultaneously introduce tunable prompts in both textual and visual encoders.
Also, different from existing related efforts~\cite{Ju:prompt-for-video} that only insert prompt vectors into the input textual sequences, we find that preparing prompts for every layer of both encoders can further close the gap to full fine-tuning.
As observed in \cref{fig:intro}, VoP achieves competitive or superior performance than other efficient tuning protocols with only \textbf{0.1\%} parameter storage.

To 
exploit essential video-specific information,
we further design three novel video prompts from different perspectives, which can seamlessly replace conventional visual prompts in VoP. 
Specifically, %
(1) \textbf{position-specific} video prompts model the information shared between frames at the same relative position.
(2) Generated \textbf{context-specific} video prompts integrate injected contextual message from the frame sequence into the intra-frame modeling.
(3) And \textbf{function-specific} video prompts adaptively assist to learn intra- or inter-frame affinities by sensing the transformation of layer functions.
By exploring video-specific prompts, \method offers a new way to transfer pre-trained foundation models to the downstream video domain.

We compare our solutions with popular tuning strategies on MSR-VTT~\cite{Xu:MSR-VTT} (both 9k and 7k splits), DiDeMo~\cite{Hendricks:DiDeMo}, ActivityNet~\cite{Heilbron:ActivityNet} and LSMDC~\cite{Rohrbach:LSMDC}.
Learning video-specific information while maintaining the pre-trained knowledge,
our video prompts deliver an average R@1 improvement of up to 4.2\% for VoP, and therefore exceed full fine-tuning by up to 1.4\% with much fewer trainable parameters.
In summary, the main contributions of our work are three-fold:
\begin{itemize}[labelsep=0.4em, leftmargin=1em,itemindent=0em]
    \vspace{-2mm}
	\item We propose the \method as a strong baseline that effectively adapts CLIP to text-video retrieval with negligible trainable parameters.
    \vspace{-2mm}
	\item To exploit video-specific information, we further develop three video prompts respectively conditioned on the frame position, frame context, and layer function.
    \vspace{-2mm}
	\item Extensive experiments on five text-video retrieval benchmarks demonstrate that various combinations of our video prompts effectively enhance VoP, outperforming full fine-tuning with much less parameter overhead.
\end{itemize}

\section{Related Work}
\label{sec:related_work}

\paragraph{Contrastive Vision-Language Pre-Training.}    %
Benefiting from large-scale visual and textual pairs collected from the Internet, learning visual representation under natural language supervision has attracted considerable attention.
As a representative example, consuming 400 million pairs of images and texts, CLIP (Contrastive Language-Image Pre-training)~\cite{Radford:CLIP} 
matches relevant image-text pairs via training two uni-modal encoders with a contrastive loss.
And the success of derivative works demonstrates the potential of adapting pre-trained vision-language models 
like CLIP~\cite{Radford:CLIP}, ALIGN~\cite{Jia:ALIGN}  %
and ALBEF~\cite{Li:ALBEF}
to various downstream applications~\cite{Patashnik:StyleCLIP,Luo:CLIP4Clip,Du:detection-prompt}.
In the video-language understanding area, some existing efforts such as HowTo100M~\cite{Miech:HowTo100M} and Frozen in Time~\cite{Bain:Frozen} have attempted to pre-train on large-scale video datasets, aiming to improve video-text representations for downstream tasks.    %
Despite the progress made, the extremely noisy text supervision of instructional videos requires a much larger scale of video-language pre-training to achieve competitive results.
In this work, we follow CLIP4Clip~\cite{Luo:CLIP4Clip} to explore the adaptation of pre-trained CLIP to the text-video retrieval task.

\vspace{-2mm}
\paragraph{Text-Video Retrieval.}

Aiming to match semantically similar samples across text and video modalities, text-video retrieval methods commonly apply a dual-branch structure to align the uni-modal features extracted by individual encoders.
While most early 
efforts
designed dedicated cross-modal fusion mechanisms after extracting offline %
features~\cite{Yu:JSFusion,Liu:multi-experts-VTR,Gabeur:MMT,Wu:HANet}, task-specific end-to-end fine-tuning from large-scale pre-trained models has recently achieved noticeable results.
For example, 
ClipBERT~\cite{Lei:ClipBERT} suggested that end-to-end fine-tuning with just a few sparsely sampled clips could outperform using densely extracted offline features from full-length videos.
CLIP4Clip~\cite{Luo:CLIP4Clip} investigated three similarity calculation mechanisms based on pre-trained CLIP, and further post-pretrained the CLIP on large-scale video-text data to improve both zero-shot and fine-tuned performance.
And X-Pool~\cite{Gorti:X-Pool} employed cross-modal attention for a text to attend its most semantically similar frames, thus generating text-conditioned video representations for retrieval.
Different from the above methods, 
we seek harmony between efficacy and parameter efficiency.
Around prompt tuning, we propose a series of solutions that reduce the overhead of trainable parameters while maintaining promising performance, thus 
decreasing
the difficulties of adaption.

\vspace{-2mm}
\paragraph{Prompt Learning.}
Stemming from 
advances in natural language processing (NLP), prompt learning initially fills the sample into properly handcrafted prompt templates, so that a pre-trained language model can ``understand'' the task~\cite{Brown:LM-fS-learners}. 
However, designing handcrafted templates requires extensive expert knowledge and limits the flexibility.
Therefore, follow-up works treat prompts as task-specific continuous vectors and directly optimize them during fine-tuning, known as \textit{prompt tuning}~\cite{Liu:prompt-tuning,Lester:prompt-tuning}.
Inspired by CLIP that embeds the textual labels of to-be-recognized objects into descriptive texts for image recognition, CoOp~\cite{Zhou:CoOp} 
applies trainable text prompts
to promote 
few-shot image classification.
As such procedure can be easily transformed into the form of various vision-language problems, text prompts have been adopted for video-understanding tasks including action recognition, action localization and text-video retrieval~\cite{Wang:ActionCLIP,Ju:prompt-for-video}.
Recently, by inserting prompt tokens into the patch token sequence or padding prompt pixels for the input image, pioneer works have successfully applied the prompt tuning method to vision backbones~\cite{Jia:VPT,Bahng:visual-prompting}.
However, whether visual prompt tuning is effective to the video domain and multi-modal applications remains untouched.
In this paper, we not only study prompt tuning for co-operative uni-model encoders, but also explore the video prompts that model a variety of video-specific information, which is the first time to our knowledge.
\section{Methodology}

\begin{figure*}[htbp]
  \centering
   \includegraphics[width=\linewidth]{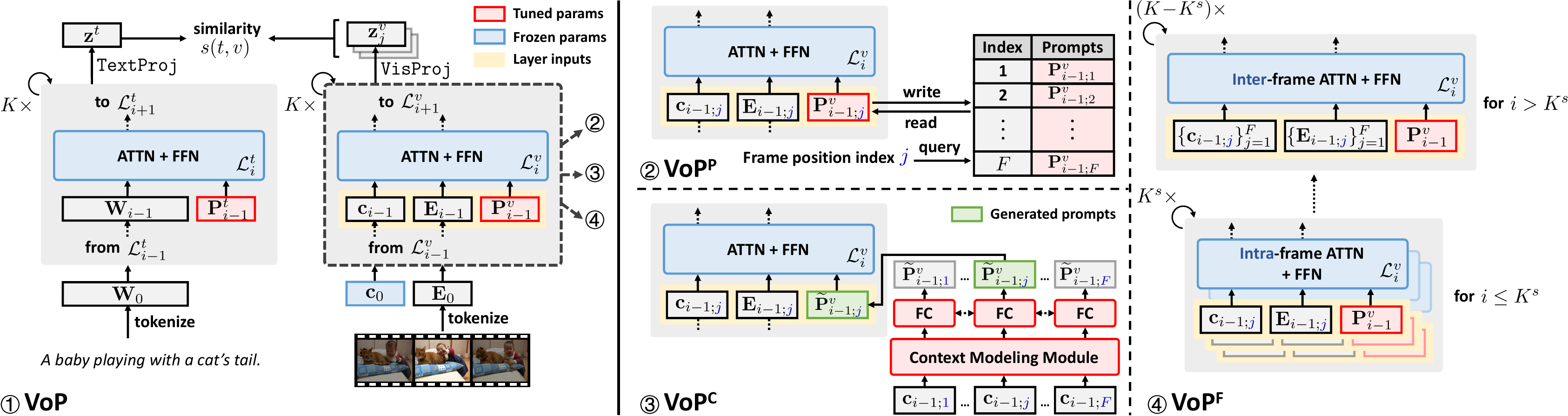}
   \vspace{-6mm}
   \caption{
   \textbf{Overview of our VoP framework before and after being equipped with video prompts}.
   To efficiently adapt CLIP to text-video retrieval, \ding{192}~VoP tunes the prompts introduced in all layers of both uni-modal encoders while keeping the rest of the model frozen.
   In addition, \ding{193}~position-specific, \ding{194}~context-specific, and \ding{195}~function-specific video prompts can replace conventional visual prompts to model essential information of the frame position, frame context, and layer function, respectively.
   }
   \label{fig:method}
   \vspace{-2mm}
\end{figure*}

In this section, we begin with a review of adapting the pre-trained CLIP to text-video retrieval (\cref{sec:preliminary}).
Then we introduce our proposed VoP 
that collaboratively promote the cross-modal alignment with negligible trainable parameters (\cref{sec:VoP}).
Finally, we further devise a series of video prompts with the consideration of the inherent nature of video, leading to superior performance than full fine-tuning (\cref{sec:video-prompt-solutions}).
The overall framework is illustrated in~\cref{fig:method}.

\subsection{Preliminary}    %
\label{sec:preliminary}

\paragraph{Problem Formulation.}
Given the text set $\mathcal{T}$ and video set $\mathcal{V}$, the objective of text-video retrieval is to learn a similarity function $s$, which produces a high similarity score $s(t, v)$ if a text $t \in \mathcal{T}$ and a video $v \in \mathcal{V}$ are semantically similar, %
while producing a low score for an irrelevant video-text pair.
Then we can rank all videos according to the query text for text-to-video retrieval (denoted as $t2v$), or rank all texts according to the query video for video-to-text retrieval (denoted as $v2t$).
In this paper, we define a text $t$ as a sequence of $N$ tokenized words, and a video $v \in \mathbb{R}^{F \times 3 \times H \times W}$ as a sequence of $F$ sampled image frames in time. %

\vspace{-2mm}
\paragraph{Revisiting CLIP-based Solution.}
Following the recent works~\cite{Luo:CLIP4Clip,Zhao:CenterCLIP,Gorti:X-Pool}, our work applies CLIP~\cite{Radford:CLIP} as the pre-trained backbone to benefit from its strong downstream potential.
Due to the large-scale contrastive image-text pre-training, 
the textual and visual encoders of CLIP share 
a joint latent space, 
where 
cross-modal
embeddings from a relevant pair can be well aligned.
Specifically, the \textbf{text} encoder first tokenizes the input text description into the word sequence, and then projects them into word embeddings $\mathbf{W}_0 = \{\mathbf{w}^1_0, \mathbf{w}^2_0, \cdots, \mathbf{w}^N_0\} \in \mathbb{R}^{N \times d^t}$.
$\mathbf{W}_0$ is fed into a $K$-layer Transformer with the architecture modifications described in BERT~\cite{Radford:BERT}, and for the $i$-th layer $\mathcal{L}^t_i$, 
\begin{align}
\mathbf{W}_i = \mathcal{L}^t_i(\mathbf{W}_{i-1})   \qquad  i=1,2,\cdots,K.   \label{eq:text-encoder}
\end{align}
And the final text embedding $\mathbf{z}^t \in \mathbb{R}^d$ is obtained by projecting the last token, which corresponds to the \texttt{[EOS]} (the end of sequence) token, from the last layer of the text encoder, \textit{i.e.}, $\mathbf{z}^t = \texttt{TextProj}(\mathbf{w}^N_K)$.
For the \textbf{visual} encoder, the input image $I$ is first split into $M$ non-overlapping patches, and projected into a sequence of patch tokens $\mathbf{E_0} \in \mathbb{R}^{M \times d^v}$.
Then, $\mathbf{E_0}$ is input into a $K$-layer Transformer-based architecture along with a learnable \texttt{[CLS]} token $\mathbf{c}_0$.
For the $i$-th layer $\mathcal{L}^v_i$, 
\begin{align}
[\mathbf{c}_i, \mathbf{E}_i] = \mathcal{L}^v_i ([\mathbf{c}_{i-1}, \mathbf{E}_{i-1}])   \qquad  i=1,2,\cdots,K,   \label{eq:vision-encoder}
\end{align}
where $[\cdot, \cdot]$ indicates concatenation on the sequence length dimension.
The final image embedding $\mathbf{z}^I \in \mathbb{R}^d$ is obtained by projecting the \texttt{[CLS]} token from the last layer of the visual encoder, \textit{i.e.}, $\mathbf{z}^I = \texttt{VisProj}(\mathbf{c}_K)$.
Therefore, the similarity score $s(t, I)$ between the image and the text can be calculated as the cosine similarity of $\mathbf{z}^t$ and $\mathbf{z}^I$.

To adapt CLIP for videos, 
a common solution is to learn a video embedding $\bar{\mathbf{z}}^v$ based on the frame embeddings $\mathbf{Z}^v = \{\mathbf{z}^v_1, \mathbf{z}^v_2, \cdots, \mathbf{z}^v_F\} \in \mathbb{R}^{F \times d}$ of all the sampled frames of $v$.
And the similarity score between $\mathbf{z}^t$ and $\bar{\mathbf{z}}^v$ can be calculated and used for retrieval.
In this paper, we focus on adapting CLIP in a parameter-efficient manner.    %
Therefore, to avoid involving extra parameters, we here apply the non-parametric approach, \textit{i.e.}, taking the average of all frame embeddings as the video embedding, as the starting point.
Note that the direction of attaching other heavy architectures on the top of CLIP~\cite{Luo:CLIP4Clip,Gorti:X-Pool} is orthogonal to our exploration, as our modifications have all occurred 
inside the encoders.
And we leave the investigation of their combination for future work.

\subsection{Text-Video Co-operative Prompt Tuning (VoP)}
\label{sec:VoP}

By keeping the backbone fixed and only optimizing the introduced trainable continuous embeddings (\textit{i.e.}, \textit{prompts}) during fine-tuning, prompt tuning~\cite{Liu:prompt-tuning,Liu:P-Tuning-deep,Bahng:visual-prompting} effectively reduces per-task storage and memory usage when adapting large-scale foundation models to downstream tasks.
In the less-studied video domain, a recent literature~\cite{Ju:prompt-for-video} proposes to insert continuous prompts into the input textual embedding sequence and shows promising results on several public video benchmarks.
However, we argue this approach remains two challenges that limit the tuning performance.
First, learning prompts only for the text branch overlooks the potential of collaboratively tuning the visual encoder.
Second, prompting the mere input layer has only a relatively indirect impact on the output embeddings.
To address the above challenges, we propose Text-Video Co-operative Prompt Tuning (VoP) that inserts prompts in each layer of both visual and text encoders (\cref{fig:method}~\ding{192}), fully excavating the knowledge embedded in the CLIP model.

Specifically, in the \textbf{text} branch, we introduce a set of learnable tokens (\textit{i.e.}, textual prompts) into each layer of the text encoder.
The textual prompts for the $i$-th layer is denoted as $\mathbf{P}^t_{i-1} \in \mathbb{R}^{P^t \times d^t}$, where $P^t$ is the number of the textual prompt tokens.
Therefore, \cref{eq:text-encoder} can be transformed as
\begin{align}
[ \underline{\quad}, \mathbf{W}_i] = \mathcal{L}^t_i([\mathbf{P}^t_{i-1}, \mathbf{W}_{i-1}]),
\end{align}
where ``$\underline{\quad}$'' indicates the output tokens at the corresponding positions will be discarded.
Similarly, in the \textbf{vision} branch, visual prompts are appended to each layer of the visual encoder.
The visual prompts for the $i$-th layer is denoted as $\mathbf{P}^v_{i-1} \in \mathbb{R}^{P^v \times d^v}$, where $P^v$ is the number of the visual prompt tokens.
And \cref{eq:vision-encoder} can be transformed as
\begin{align}
[\mathbf{c}_i, \underline{\quad}, \mathbf{E}_i] = \mathcal{L}^v_i ([\mathbf{c}_{i-1}, \mathbf{P}^v_{i-1}, \mathbf{E}_{i-1}]).    \label{eq:visual-prompts}
\end{align}

By jointly minimizing the symmetric text-to-video and video-to-text cross-entropy losses~\cite{Luo:CLIP4Clip}, both textual and visual prompts are fine-tuned while the other parameters from the two encoders are frozen.
We find that the co-operation of entirely prompting both encoders adequately adapts the latent space of the model to the target domain.

\subsection{Equipping with Video Prompts}
\label{sec:video-prompt-solutions}

Being a mature and efficient solution for text-video retrieval, however, the current VoP faces the dilemma that it treats frames as independent images, making 
it 
difficult to utilize rich information other than single-frame content.
Therefore, we further develop a series of video prompts (\cref{fig:method}~\ding{193}~\ding{194}~\ding{195}) specifically for processing videos, which excavates information from different perspectives.
We describe how these video prompts are combined with VoP as follows.

\vspace{-4mm}
\paragraph{VoP with Position-Specific Video Prompts (VoP$^\text{P}$).}
One shortcoming of the current VoP is the learned prompts are shared for all frames, ignoring the order of the frame sequence.
To inject information about the relative position of the current input frame, we present position-specific video prompts (\cref{fig:method}~\ding{193} VoP$^\text{P}$), where visual prompts are only allowed to be shared between all frames at the same relative position in their belonged videos.
This can be implemented by maintaining a table, where keys are the position indices and values are the prompts.
Thus, the prompts for the current frame can be read by querying the table with the frame position index, and be written back after their optimization.
Formally, after introducing the frame position, the flow of each visual encoder layer in \cref{eq:visual-prompts} now becomes %
\begin{align}
[\mathbf{c}_{i; \textcolor{blue}{j}}, \underline{\quad}, \mathbf{E}_{i; \textcolor{blue}{j}}] = \mathcal{L}^v_i ([\mathbf{c}_{i-1; \textcolor{blue}{j}}, \mathbf{P}^v_{i-1; \textcolor{blue}{j}}, \mathbf{E}_{i-1; \textcolor{blue}{j}}]),
\end{align}
where $j$ is the position index of the current frame in the video, and $\mathbf{P}^v_{i-1;j} \in \mathbb{R}^{P^v \times d^v}$ is the visual prompts of the $i$-th layer shared for all the $j$-th frames in all videos.
Allowing to have more tunable position-specific parameters, VoP$^\text{P}$ increases the capacity for informative videos.
In practice, as a copy of prompts for each layer contains several prompt tokens, we found that changing not all, but only a part of position-agnostic tokens into position-specific ones is sufficient for both effectiveness and efficiency.

\vspace{-4mm}
\paragraph{VoP with Context-Specific Video Prompts (VoP$^\text{C}$).}
Another piece of information that cannot be exploited by the current 
scheme is the contextual relationships in videos.
When addressing the current frame, a natural intuition is to integrate the contextual information from the rest of the video to emphasize important elements.
To convert such information into prompts, we propose the dynamically generated context-specific video prompts (\cref{fig:method}~\ding{194} VoP$^\text{C}$) that are input-conditional rather than fixed once learned.
Specifically, in each layer, we form the \texttt{[CLS]} tokens of frames from the same video into a sequence $\mathbf{C}_{i-1} \in \mathbb{R}^{F \times d^v}$, \textit{i.e.}, 
$\mathbf{C}_{i-1} = \{ \mathbf{c}_{{i-1}; \textcolor{blue}{1}}, \mathbf{c}_{{i-1}; \textcolor{blue}{2}}, \cdots, \mathbf{c}_{{i-1}; \textcolor{blue}{j}} \cdots, \mathbf{c}_{{i-1}; \textcolor{blue}{F}} \}$,
and feed $\mathbf{C}_{i-1}$ into a Context Modeling Module (CMM) to modulate each frame token with its contextual information:
\begin{align}
\widetilde{\mathbf{C}}_{i-1} = \texttt{CMM}(\mathbf{C}_{i-1}).
\end{align}

Then, a fully-connected (FC) layer generates the prompt tokens conditioned on the modulated frame token, where the input vector is stretched by the projection before splitting into multiple tokens:
\begin{align}
\widetilde{\mathbf{P}}^v_{i-1; \textcolor{blue}{j}} = \texttt{FC}(\tilde{\mathbf{c}}_{i-1; \textcolor{blue}{j}}).
\end{align}
Therefore, \cref{eq:visual-prompts} now becomes
\begin{align}
[\mathbf{c}_{i; \textcolor{blue}{j}}, \underline{\quad}, \mathbf{E}_{i; \textcolor{blue}{j}}] = \mathcal{L}^v_i ([\mathbf{c}_{i-1; \textcolor{blue}{j}}, \widetilde{\mathbf{P}}^v_{i-1; \textcolor{blue}{j}}, \mathbf{E}_{i-1; \textcolor{blue}{j}}]).
\end{align}

In this way, global contextual information can be encoded into the generated prompts and participate in the intra-frame modeling.
Note that $\widetilde{\mathbf{C}}_{i-1}$ does not pass through the next layer since it is used for prompt generation.
To avoid introducing excessive parameters to be trained, parameters of CMM and the FC layer are shared between all encoder layers.
As there exist several options for the architecture of CMM, experiment results show that the Bi-directional Long Short-Term Memory (BiLSTM)~\cite{Graves:bi-LSTM} is an overall preferable choice.

\vspace{-4mm}
\paragraph{VoP with Function-Specific Video Prompts (VoP$^\text{F}$).}
Although we have designed two special prompts in consideration of the inherent properties of video, they are hardly a complete substitute for spatio-temporal modeling, which leads to the outstanding performance of video Transformers~\cite{Bertasius:TimeSformer}.
However, attaching even a ``lightweight'' Transformer on top of the CLIP visual encoder will increase the number of training parameters by a significant amount.
To obtain a ``free'' spatio-temporal modeling, we propose a transformation of the functionality of existing frozen parameters in the deeper layers (\cref{fig:method}~\ding{195} VoP$^\text{F}$).
Specifically, we split the current visual encoder into two parts according to the depth of the layer, and each part undertakes different functions.
The first part that contains $K^s$ shallow layers still performs spatial self-attention for tokens of each frame, and video prompts discussed above can still be adopted in this part without changes.
However, the second part that contains the last $(K\!-\!K^{s})$ layers now performs inter-frame spatio-temporal self-attention without changing structure.
And visual prompts in these layers are prepared for the input sequence that consists of \texttt{[CLS]} and patch tokens from all frames of the same video.
In other words, following the change of functions at different layers, the visual prompts are adaptively divided into frame-level and video-level ones.
Formally, for the $i$-th layer $\mathcal{L}^v_i$, \cref{eq:visual-prompts} remains unchanged when $i\!\leq\!K^s$.
And for $i\!>\!K^s$,
\cref{eq:visual-prompts} becomes
\begin{flalign}
&[\mathbf{C}_i, \underline{\quad}, \mathbf{E}_{i;\textcolor{blue}{1}}, \mathbf{E}_{i;\textcolor{blue}{2}}, \cdots, \mathbf{E}_{i;\textcolor{blue}{F}}] \nonumber  \\
&= \mathcal{L}^v_i ([\mathbf{C}_{i-1}, \mathbf{P}^v_{i-1}, \mathbf{E}_{i-1;\textcolor{blue}{1}}, \mathbf{E}_{i-1;\textcolor{blue}{2}}, \cdots, \mathbf{E}_{i-1;\textcolor{blue}{F}}]).
\end{flalign}

We note that before feeding into the $(K^s\!+\!1)$-th layer, a trainable frame positional embedding is added to all tokens from the video to retain positional information, which is omitted in the formula for simplification.
And we use VoP$^\text{F+P}$ and VoP$^\text{F+C}$ to indicate the deployment of position-specific and context-specific video prompts at $K^s$ shallow layers while applying VoP$^\text{F}$, respectively.

\section{Experiments}

\begin{table*}[!t]    %
\tablestyle{5pt}{1.0}
\setlength\tabcolsep{4pt}
\def\w{20pt} 
\scalebox{1}{
    \begin{tabular}{lc|cccp{0.7cm}<{\centering}p{0.85cm}<{\centering}|cccp{0.7cm}<{\centering}p{0.75cm}<{\centering}}
    \multirow{2}[1]{*}{\textbf{Methods}} & \multirow{2}[1]{*}{\textbf{Params (M)}} & \multicolumn{5}{c}{\textbf{$t2v$}}     & \multicolumn{5}{c}{\textbf{$v2t$}} \\
          &       & \textbf{R@1} & \textbf{R@5} & \textbf{R@10} & \textbf{MnR↓} & \textbf{MdR↓} & \textbf{R@1} & \textbf{R@5} & \textbf{R@10} & \textbf{MnR↓} & \textbf{MdR↓} \\
    \shline
    Full  & 119.8 (100\%) & 41.7  & 69.2  & 79.0  & 16.5  & 2.0   & 42.5  & \uline{70.9}  & \textbf{81.4} & \textbf{11.0} & 2.0    \\
    Bias~\cite{Cai:ft-bias}  & 0.1 (0.104\%) & 39.7  & 66.5  & 77.3  & 17.3  & 2.0   & 41.1  & 68.4  & 79.2  & 13.6  & 2.0   \\
    Proj~\cite{Jia:VPT}  & 0.7 (0.547\%) & 37.1  & 63.0  & 76.1  & 20.5  & 3.0   & 37.2  & 64.6  & 75.9  & 16.7  & 3.0   \\
    Partial~\cite{Jia:VPT} & 7.7 (6.410\%) & 39.8  & 65.3  & 75.9  & 19.3  & 2.0   & 37.9  & 66.1  & 77.4  & 15.5  & 3.0   \\
    Adapter$^\text{ATTN}$~\cite{He:unify-PET} & 2.0 (1.655\%) & 37.6  & 63.2  & 75.8  & 18.7  & 3.0   & 39.6  & 66.5  & 76.8  & 14.7  & 2.0   \\
    Adapter$^\text{FFN}$~\cite{Chen:AdaptFormer} & 2.0 (1.655\%) & 38.2  & 63.5  & 76.4  & 17.9  & 3.0   & 39.9  & 66.8  & 77.7  & 14.2  & 2.0   \\
    \hline
    \textbf{VoP} & 0.1 (0.103\%) & 39.6  & 66.7  & 77.8  & 17.2  & 2.0   & 42.1  & 68.8  & 80.7  & 12.4  & 2.0   \\
    \textbf{VoP$^\text{P}$} & 0.5 (0.441\%) & 40.1  & 65.7  & 77.7  & 16.9  & 2.0   & 42.5  & 70.0  & 79.9  & 12.4  & 2.0   \\
    \textbf{VoP$^\text{C}$} & 14.3 (11.898\%) & 40.8  & 68.1  & 79.0  & \uline{15.8}  & 2.0   & 42.3  & 70.1  & 81.1  & \uline{11.4}  & 2.0   \\
    \textbf{VoP$^\text{F}$} & 0.1 (0.103\%) & 42.6  & 68.4  & 78.7  & \uline{15.8}  & 2.0   & 42.4  & 70.5  & 81.0  & \textbf{11.0} & 2.0   \\
    \hline
    \specialrule{0em}{0.5pt}{0.5pt}
    \textbf{VoP$^\text{F+P}$} & 0.4 (0.328\%) & \uline{43.5}  & \uline{69.3}  & \uline{79.3}  & \textbf{14.8} & 2.0   & \uline{43.6}  & \textbf{71.2} & \uline{81.2}  & \textbf{11.0} & 2.0   \\
    \textbf{VoP$^\text{F+C}$} & 14.1 (11.785\%) & \textbf{44.6} & \textbf{69.9} & \textbf{80.3} & 16.3  & 2.0   & \textbf{44.5} & 70.7  & 80.6  & 11.5  & 2.0   \\
    \end{tabular}%
    }
    \vspace{-2mm}
  \caption{\textbf{Retrieval results on the MSR-VTT-9k dataset}. 
  }
  \label{tab:main_results_MSRVTT9k}%
  \vspace{-3mm}
\end{table*}%

\subsection{Experimental Setup}

\paragraph{Datasets.} \label{dataset_intro}
We conduct our experiments on the following benchmarks for text-video retrieval:
(1) \textbf{MSR-VTT}~\cite{Xu:MSR-VTT} contains 10,000 videos, each paired with about 20 captions.
Following previous works~\cite{Luo:CLIP4Clip,Zhao:CenterCLIP,Gorti:X-Pool}, we report results on %
both two data splits, `training-9K'~\cite{Gabeur:MMT} and `training-7K'~\cite{Miech:MSRVTT-7k}, to compare with baselines.
The test data in both splits is `test 1k-A', which is comprised of 1,000 video-text pairs following JSFusion~\cite{Yu:JSFusion}.
We use `MSR-VTT-9k' and `MSR-VTT-7k' to refer to the two data splits, respectively.
(2) \textbf{DiDeMo}~\cite{Hendricks:DiDeMo} contains 10,000 Flickr videos with 40,000 sentences.
Following the setting from \cite{Liu:multi-experts-VTR,Lei:ClipBERT,Bain:Frozen}, we concatenate all the sentences of a video to form a paragraph and evaluate the model with paragraph-video retrieval.
(3) \textbf{ActivityNet}~\cite{Heilbron:ActivityNet} contains 20,000 YouTube videos.
Following \cite{Gabeur:MMT,Luo:CLIP4Clip}, all descriptions of a video are also concatenated into a single query, and the `val1' split is used to evaluate the model.
(4) \textbf{LSMDC}~\cite{Rohrbach:LSMDC} contains 118,081 video clips extracted from 202 movies.
There are 109,673 videos in the training set and 7,408 videos in the validation set.
And 1,000 videos in the test set are from movies disjoint with the training and validation set.

\vspace{-4mm}
\paragraph{Evaluation Metrics.}
We follow the standard retrieval metrics~\cite{Luo:CLIP4Clip} to use R@K (recall at rank K, higher is better), MnR (mean rank, lower is better), and MdR (median rank, lower is better) for evaluation.
Specifically, R@1, R@5, and R@10 are reported.    %

\vspace{-4mm}
\paragraph{Baselines.}
We compare our methods with other commonly used fine-tuning protocols:
(1) \textbf{Full}: fully update all parameters of the pre-trained backbone.
(2) \textbf{Bias}~\cite{Cai:ft-bias,Zaken:ft-bias}: fine-tune only the bias terms of the pre-trained backbone.
(3) \textbf{Proj}~\cite{Jia:VPT}: fine-tune only the last linear projection of both encoders.
(4) \textbf{Partial}~\cite{Jia:VPT}: fine-tune only the last layer of both encoders.
(5) \textbf{Adapter$^\text{ATTN}$}~\cite{Houlsby:Adapter,He:unify-PET}: fine-tune only the FC layers inserted in parallel to each multi-head self-attention layer in both encoders.
(6) \textbf{Adapter$^\text{FFN}$}~\cite{Houlsby:Adapter,Chen:AdaptFormer}: fine-tune only the FC layers inserted in parallel to each feed-forward network in both encoders.

\vspace{-2mm}
\paragraph{Implementation Details.}
12-layer visual and text encoders are adopted from a pre-trained CLIP (ViT-B/32+Transformer),
and all original parameters of the backbone are kept frozen unless otherwise stated.
We optimize each model for 5 epochs using the AdamW~\cite{Loshchilov:AdamW} optimizer with weight decay set to 0.2, and decay the learning rate using a cosine schedule~\cite{Bjorck:cosine-decay}.
And the initial learning rate for each method is determined by searching in the range of [$1e^{-6}$, $1e^{-2}$]. %
For all experiments, we uniformly sample 12 frames from each video following previous studies~\cite{Bain:Frozen,Luo:CLIP4Clip,Gorti:X-Pool}.
All video frames are resized to 224$\times$224, and the maximum number of textual tokens is 77, following the original CLIP design.
And the batch size is set to 32.
For Adapter$^\text{ATTN}$ and Adapter$^\text{FFN}$, the number of hidden dimensions is set to 64. 
For methods with VoP, the prompt length for both encoders, \textit{i.e.}, $P^t$ and $P^v$, are set to 8 as default.
And a normal initialization is applied to prompt parameters.
For VoP$^\text{P}$, VoP$^\text{C}$, VoP$^\text{F+P}$ and VoP$^\text{F+C}$, the length of video prompts is set to 4.
And context-specific video prompts applies a 1-layer BiLSTM~\cite{Graves:bi-LSTM} as CMM to pass contextual information.
For VoP$^\text{F}$, VoP$^\text{F+P}$ and VoP$^\text{F+C}$, we set $K^s$ as 8.
All experiments are carried out on 4 NVIDIA Tesla V100 GPUs.

\begin{figure}[!t]    %
  \centering
   \includegraphics[width=\linewidth]{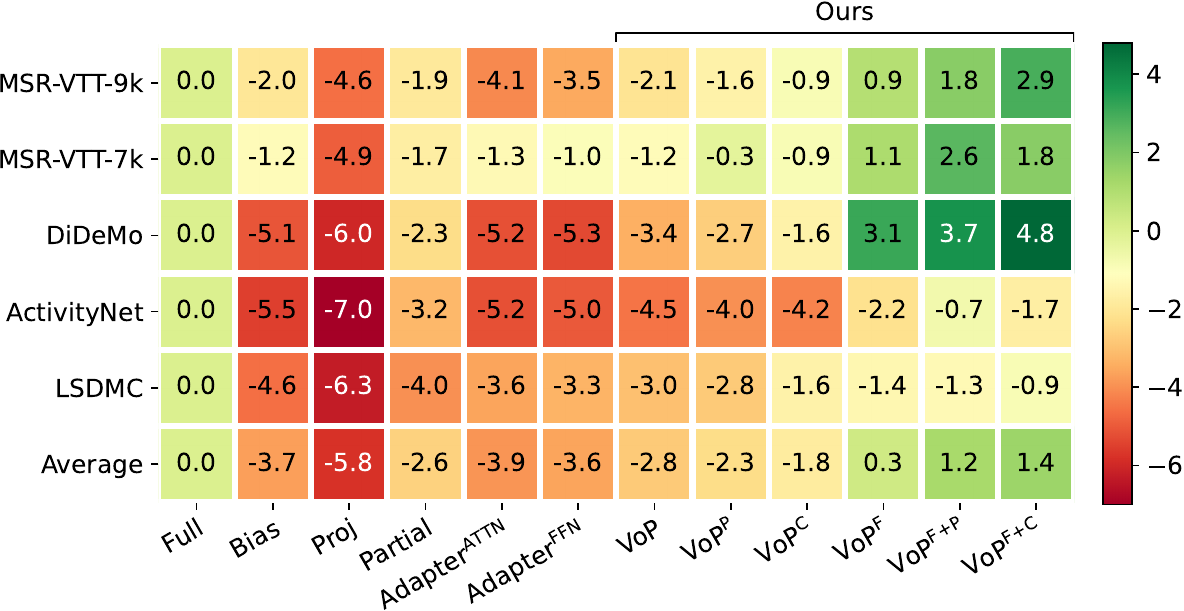}
   \vspace{-6mm}
   \caption{
   \textbf{$t2v$ R@1 gains of all methods in comparison against Full}. The first five rows show the improvement or deterioration on each benchmark, and the last row summarizes the average relative results over all benchmarks.
   }
   \label{fig:R1_vs_Full_heatmap}
   \vspace{-2mm}
\end{figure}

\subsection{Main Results}    \label{sec:main_results}

We compare our methods with popular tuning protocols on five benchmarks.
We here represent $t2v$ R@1 gains relative to full fine-tuning in \cref{fig:R1_vs_Full_heatmap}, and report $t2v$ and $v2t$ results on MSR-VTT-9k in \cref{tab:main_results_MSRVTT9k}.
Detailed results on the other four benchmarks can be found in the supplementary materials.
We here highlight some important observations from \cref{fig:R1_vs_Full_heatmap} as follows:
\begin{itemize}[labelsep=0.4em, leftmargin=1em,itemindent=0em]
    \vspace{-2mm}
	\item \textbf{VoP achieves competitive or superior performance than other efficient tuning protocols with only 0.1\% parameter storage}. 
The only exception is that R@1 of Partial is on average 0.2\% higher than that of our VoP at the expense of a more than 60$\times$ parameter overhead.
    \vspace{-2mm}
    \item \textbf{Appending only one video prompts can significantly improve VoP to even compete with Full}.    %
Position-specific, context-specific and function-specific video prompts respectively bring 0.5\%, 1.0\% and 3.1\% average R@1 gain.
While VoP$^\text{P}$ and VoP$^\text{C}$ obtain a superior performance than all other efficient tuning protocols on average, VoP$^\text{F}$ even outperforms Full by 0.3\% without introducing more parameters to VoP.
    \vspace{-2mm}
    \item \textbf{Combining the video prompts can lead to higher performance benefits}. %
Compared to VoP$^\text{F}$, VoP$^\text{F+P}$ and VoP$^\text{F+C}$ further achieve 0.9\% and 1.1\% R@1 improvements.
As a conclusion, we provide sufficient solutions to choose from according to the strictness of the parameter and computation limitations.
\end{itemize}

\begin{table}[!t]
\tablestyle{5pt}{1.0}
\setlength\tabcolsep{4pt}
\def\w{20pt} 
\scalebox{1}{
    \begin{tabular}{p{0.8cm}<{\centering}p{0.8cm}<{\centering}|ccc|p{0.6cm}<{\centering}p{0.6cm}<{\centering}}
    \textbf{Textual} & \textbf{Visual} & \textbf{R@1} & \textbf{R@5} & \textbf{R@10} & \textbf{MnR↓} & \textbf{MdR↓} \\
    \shline
          &       & 31.5  & 52.8  & 63.6  & 42.9  & 5.0   \\
          & \cmark & 36.5  & 62.7  & 75.1  & 18.3  & 3.0   \\
    \cmark &       & 36.3  & 63.4  & 75.0  & 20.3  & 3.0   \\
    \cmark & \cmark & \textbf{39.6} & \textbf{66.7} & \textbf{77.8} & \textbf{17.2} & \textbf{2.0} \\
    \end{tabular}%
    }
  \vspace{-2mm}
  \caption{\textbf{Ablation on co-operative multi-modal prompts in VoP.} The co-operation of prompting multi-modal branches leads to higher performance.}
  \label{tab:VoP-ablation}%
  \vspace{-2mm}
\end{table}%

\begin{table*}[!t]
\tablestyle{5pt}{1.0}
\setlength\tabcolsep{4pt}
\def\w{20pt} 
\scalebox{1}{
    \begin{tabular}{l|ccc|ccc|ccc|ccc|ccc}
    \multirow{2}[0]{*}{\textbf{Choice of CMM}} & \multicolumn{3}{c}{\textbf{MSR-VTT-9k}} & \multicolumn{3}{c}{\textbf{MSR-VTT-7k}} & \multicolumn{3}{c}{\textbf{DiDeMo}} & \multicolumn{3}{c}{\textbf{ActivityNet}} & \multicolumn{3}{c}{\textbf{LSMDC}} \\
          & \textbf{R@1} & \textbf{R@5} & \textbf{R@10} & \textbf{R@1} & \textbf{R@5} & \textbf{R@10} & \textbf{R@1} & \textbf{R@5} & \textbf{R@10} & \textbf{R@1} & \textbf{R@5} & \textbf{R@10} & \textbf{R@1} & \textbf{R@5} & \textbf{R@10} \\
    \shline
    Transformer & 40.1  & 68.2  & 78.8  & 39.5  & 68.2  & 78.1  & \textbf{40.4} & 67.3  & 77.3  & 32.0  & 61.5  & 74.9  & 20.3  & 39.5  & 47.8  \\
    LSTM  & 40.6  & \textbf{69.5} & \textbf{79.7} & 39.5  & \textbf{69.3} & 78.0  & 38.6  & 66.7  & 77.0  & 32.4  & 62.0  & 75.4  & 19.6  & 38.2  & 47.7  \\
    BiLSTM & \textbf{40.8} & 68.1  & 79.0  & \textbf{40.0} & 67.3  & \textbf{78.2} & 40.0  & \textbf{68.0} & \textbf{78.5} & \textbf{32.6} & \textbf{62.5} & \textbf{76.5} & \textbf{20.4} & \textbf{40.0} & \textbf{48.1} \\
    \end{tabular}%
  }
  \vspace{-2mm}
    \caption{
   \textbf{Ablation on the CMM choices in VoP$^\text{C}$}. 
   We report the R@1, R@5 and R@10 results on five benchmarks.
   In general, BiLSTM achieves the best results on more datasets.}  %
  \label{tab:CSVP_CMM_ablation}%
  \vspace{-2mm}
\end{table*}%

\subsection{Ablation Study}

We ablate different model design choices on MSR-VTT-9k and report $t2v$ results if no otherwise specified.
We note that as many hyper-parameters exist in multiple solutions, we determine the values to be taken based on their performance in the base solution to speed up the search.

\vspace{-2mm}
\paragraph{Effect of Co-operative Uni-Modal Prompts in VoP.}
In \cref{tab:VoP-ablation}, we report the performance of using CLIP without fine-tuning, tuning with uni-modal prompts, and using our proposed VoP.
While uni-modal prompts outperform directly applying CLIP without tuning, our VoP achieves better results, which demonstrates that prompting both encoders enables better adaptation to the downstream text-video retrieval task.

\begin{figure}[!t]
  \centering
  \includegraphics[width=\linewidth]{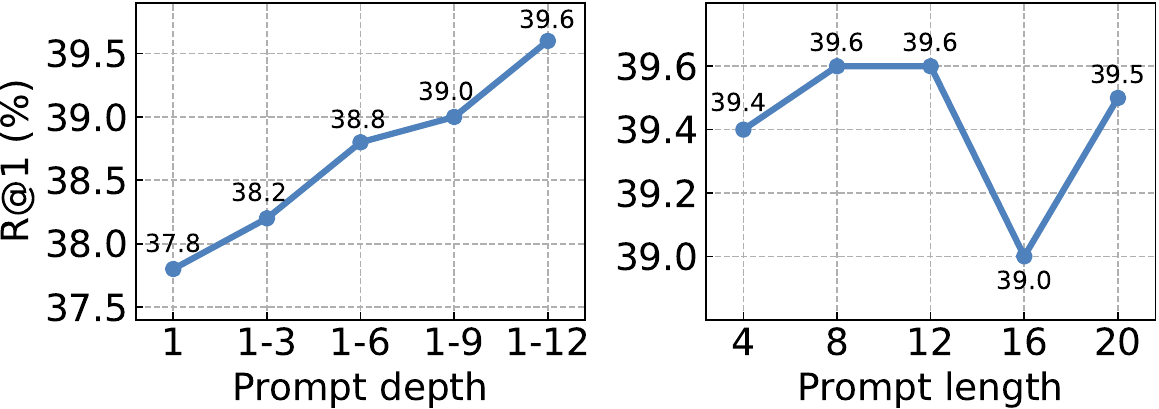}
  \vspace{-6mm}
  \caption{
  \textbf{Ablation on hyperparameters of prompts in VoP}. %
  \textbf{Left}: Ablation on prompt depth, \textit{i.e.}, where and how many layers to insert prompts. $i$-$j$ indicates the encoder layer indices that prompts are inserted into, while the 1-st layer refers to the one closest to the input. 
  \textbf{Right}: Ablation on prompt length, \textit{i.e.}, the number of prompt tokens in each layer.}
  \label{fig:VoP_prompt_depth_length}
\end{figure}

\vspace{-2mm}
\paragraph{Effect of Prompt Depth in VoP.}
In \cref{fig:VoP_prompt_depth_length} (left), we gradually increase the number of layers inserted into prompts.
In general, the performance of VoP is positively correlated with the prompt depth.
And inserting prompts into every layer of both encoders contributes to the best results.

\vspace{-2mm}
\paragraph{Effect of Prompt Length in VoP.}
In \cref{fig:VoP_prompt_depth_length} (right), we vary the number of prompt tokens in each layer.
Unlike the prompt depth, steadily increasing the prompt length does not lead to continuous growth of performance.
And using only 8 tokens remains a competitive performance with parameter efficiency.

\vspace{-2mm}
\paragraph{Effect of Prompt Length for Video Prompts.}

\begin{figure}[!t] %
  \centering
  \includegraphics[width=\linewidth]{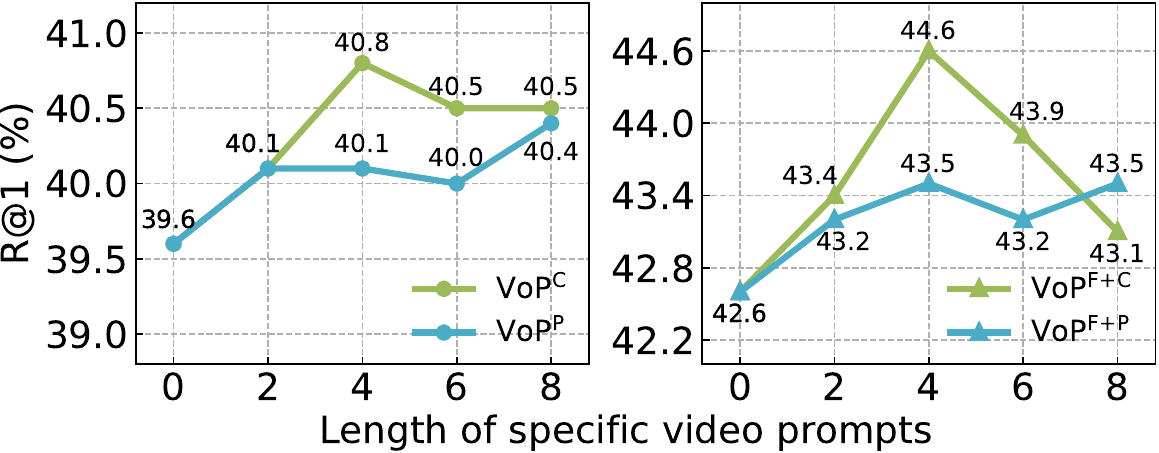}
  \vspace{-6mm}
  \caption{
  \textbf{Ablation on length of video prompts}.
  The number of visual prompt tokens that are replaced with corresponding video-specific prompts is varied, and the maximum length is set to 8 based on previous experimental results.
  }
  \label{fig:specific_prompt_length}
  \vspace{-2mm}
\end{figure}

\cref{fig:specific_prompt_length} ablates how many conventional visual prompt tokens to change to video-specific ones.
We observe that only inserting video-specific prompts does not lead to the best results for all cases.
And turning only half of the 8 prompt tokens into specific ones is a more universal choice that achieves a trade-off between effectiveness and efficiency.
We explain that the conventional visual prompts learn general knowledge shared between all frames across videos, which complements our video prompts that may focus more on other information around a specific frame.

\begin{table}[!t]
\tablestyle{5pt}{1.0}
\setlength\tabcolsep{4pt}
\def\w{20pt} 
\scalebox{1}{
  \begin{tabular}{l|ccc|p{0.6cm}<{\centering}p{0.6cm}<{\centering}}
    \textbf{Role of CMM} & \textbf{R@1} & \textbf{R@5} & \textbf{R@10} & \textbf{MnR↓} & \textbf{MdR↓} \\
    \shline
    Updating \texttt{[CLS]} & 38.0  & 63.6  & 75.3  & 18.5  & \textbf{2.0}   \\
    Generating prompts  & \textbf{40.8}  & \textbf{68.1}    & \textbf{79.0}  & \textbf{15.8} & \textbf{2.0}   \\
    \end{tabular}%
    }
    \vspace{-2mm}
    \caption{\textbf{Ablation on the role of CMM in VoP$^\text{C}$}. 
    The current VoP$^\text{C}$ design represented in the second row achieves better results.
    }
  \label{tab:CSVP_if_prompts_needed}%
\end{table}%

\begin{figure}[!t]
  \centering
  \includegraphics[width=0.65\linewidth]{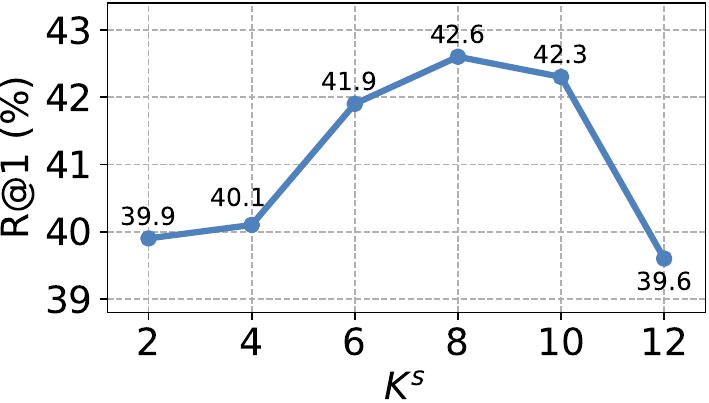}
  \vspace{-2mm}
  \caption{
  \textbf{Ablation on layers with different functions in VoP$^\text{F}$}.
  The first $K^s$ shallow layers perform intra-frame spatial self-attention, and the subsequent deep layers perform inter-frame spatio-temporal self-attention.
  }
  \label{fig:FSVP_Ks_value}
  \vspace{-2mm}
\end{figure}

\begin{figure*}[!t]    %
  \centering
   \includegraphics[width=\linewidth]{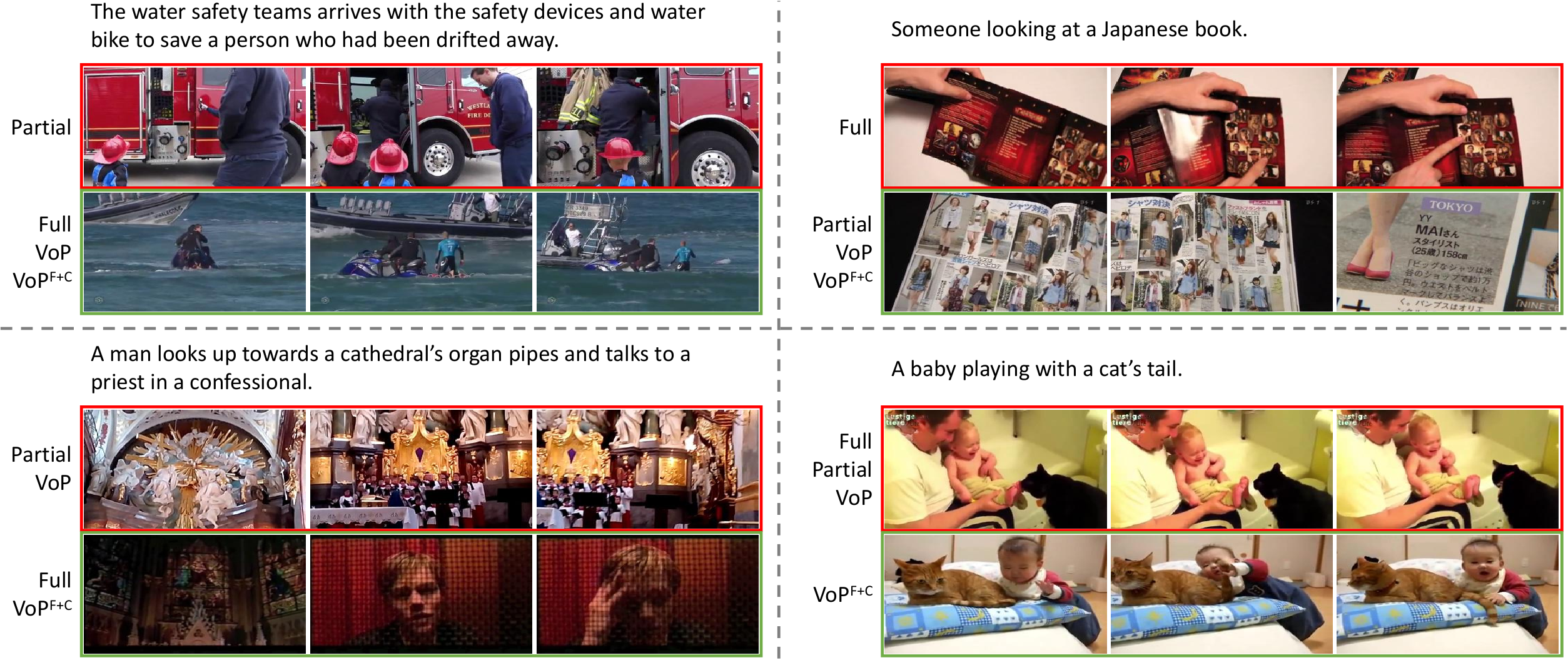}
   \vspace{-6mm}
   \caption{
   \textbf{Qualitative results of four tuning methods: Full, Partial, VoP and VoP$^\text{F+C}$}.
           Given the query text, we represent the rank-1 retrieval result of each method, which can be \textcolor{red}{incorrect} (each first row) or \textcolor{gdcolor}{ground truth} (each second row).
           }
   \label{fig:qualitative_results}
   \vspace{-4mm}
\end{figure*}

\vspace{-2mm}
\paragraph{Effect of CMM in VoP$^\text{C}$.}

We first ablate different choices on which CMM to model the contextual information.
We report the R@1 of three candidates, \textit{i.e.}, Transformer, LSTM and BiLSTM, on the five benchmarks.
Note that we use a 4-layer Transformer, which is a common choice when modeling temporal dependencies in previous works~\cite{Luo:CLIP4Clip}.
And the number of layers is set to 1 for both LSTM, and BiLSTM.
As shown in \cref{tab:CSVP_CMM_ablation}, BiLSTM is a more effective choice in general.

In VoP$^\text{C}$, unlike the common practice in temporal modeling, CMM outputs frame tokens modulated with contextual information for generating prompts %
instead of directly updating \texttt{[CLS]} tokens.
Thus
a question naturally arises: Do we indeed need to generate prompts?
In other words, will %
directly updating \texttt{[CLS]} tokens with CMM be a more effectual option?
To answer the question, we compare the two solutions in \cref{tab:CSVP_if_prompts_needed}, and our VoP$^\text{C}$ achieves leadership in nearly all metrics.
Our explanation for the results is that %
generating prompts for the token sequence achieves cross-frame communication with less disruption to the information flow of the original visual encoder.

\vspace{-2mm}
\paragraph{Effect of Split Layer in VoP$^\text{F}$.}
We vary the value of $K_s$ in VoP$^\text{F}$ and illustrate the results in \cref{fig:FSVP_Ks_value}. 
A larger value of $K_s$ means that more shallow layers are used for intra-frame spatial self-attention and fewer deep layers for inter-frame spatio-temporal self-attention.
We note that when $K_s=12$, VoP$^\text{F}$ degrades to VoP.
And the result of $K_s=0$, \textit{i.e.}, all layers performing inter-frame spatio-temporal self-attention, is not represented as we found it failed to generalize and the performance collapsed.
As shown in the figure, R@1 raises with increasing $K_s$ until $K_s=8$, which shows the necessity of intra-frame message exchanging in shallow layers.
Subsequently, R@1 begins to drop as $K_s$ continues to increase, indicating that properly substituting the functions of layers and corresponding prompts brings improvements.
And, in conclusion, $K_s=8$ is the choice that achieves the best trade-off.

\subsection{Qualitative Results}

In \cref{fig:qualitative_results}, we visualize some $t2v$ retrieval examples from the test set of MSR-VTT-9k.
We represent the retrieval results of four tuning methods: Full, Partial, VoP, and VoP$^\text{F+C}$.
In the \textbf{top left} example, Full and our proposed methods can retrieve the correct video while Partial matches an unrelated one, which shows the inferiority of existing efficient tuning protocols.
In the \textbf{top right} example, Full fails to recognize a ``Japanese'' book while parameter-efficient tuning methods succeed by capturing visual clues of Japanese characters and related English words like ``Tokyo'', indicating that updating all parameters might be an unsatisfactory strategy as more knowledge from large-scale text-image pre-training is forgotten. 
In the \textbf{bottom left} example, by fine-tuning all parameters with video datasets or designing specialized prompting solutions for videos, Full and VoP$^\text{F+C}$ can understand the whole event represented by sequenced frames. 
Even if some textual elements like ``priest'' are not visually present, the methods overcome such minor semantic misalignments and select more relevant candidates from a global view.
Finally in the \textbf{bottom right} example, understanding the concept of ``tail'' and capturing the interaction of ``playing with'', VoP$^\text{F+C}$ can distinguish the correct video from hard negative candidates while all the other three methods fail.

\section{Conclusion}

In this paper, we continue the vein of 
prompt tuning
to transfer pre-trained CLIP for text-video retrieval with both effectiveness and efficiency.
We first devise a simple but competitive baseline VoP, which achieves promising performance with only 0.1\% trainable parameters by prompting all layers of both textual and visual encoders.
To increase the revenue of VoP, we further explore three video prompts to model different video-specific information.
Different combinations of our video prompts can be selected depending on the strictness of the limits on parameter overhead, and achieve at most 1.4\% average relative improvement with much fewer trainable parameters compared to full fine-tuning.
We hope our work can inspire future research on how to fully exploit the large foundation models in challenging video-understanding tasks.

\section*{Acknowledgement}

This work was supported by STI 2030—Major Projects (2022ZD0208800), NSFC General Program (Grant No. 62176215).
This work was supported by Alibaba Group through Alibaba Research Intern Program.

\clearpage
\normalem
{\small
\bibliographystyle{ieee_fullname}
\bibliography{ref}
}

\clearpage

\appendix

\section*{Appendix}

\section{Discussion on Non-parameter-efficient Methods}

\begin{table}[th]
\tablestyle{5pt}{1.0}
\setlength\tabcolsep{2.5pt}
\def\w{20pt} 
\scalebox{1}{
  \begin{tabular}{lc|cccp{0.6cm}<{\centering}p{0.75cm}<{\centering}}
    \textbf{Methods} & \textbf{Params (M)} & \textbf{R@1} & \textbf{R@5} & \textbf{R@10} & \textbf{MnR↓} & \textbf{MdR↓} \\
    \shline
    X-Pool~\cite{Gorti:X-Pool} & 1.3 (1.1\%) & 40.5  & 64.8  & 75.0  & 18.9  & 2.0   \\
    \textbf{VoP$^\text{F}$} & \textbf{0.1 (0.1\%)} & 42.6  & 68.4  & 78.7  & 15.8  & 2.0   \\
    X-Pool~\cite{Gorti:X-Pool}+\textbf{VoP} & 1.4 (1.2\%) & \textbf{43.1}  & \textbf{69.5}  & \textbf{79.5}  & \textbf{14.5}  & 2.0   \\
  \end{tabular}
  }
  \vspace{-2mm}
  \caption{\textbf{Comparison with non-parameter-efficient X-Pool~\cite{Gorti:X-Pool} after freezing the CLIP backbone}.
  The $t2v$ retrieval results are obtained on the MSR-VTT-9k dataset.
  }
  \label{tab:results_freeze_CLIP}%
  \vspace{-2mm}
\end{table}

Our work aims to greatly reduce the overall storage costs while achieving promising cross-modal retrieval performance.
Related non-parameter-efficient methods~\cite{Gorti:X-Pool,Ma:X-CLIP,Liu:TS2-Net} requires to fine-tune the additional parameters together with the CLIP backbone, which results in an unaffordable overhead.
Despite the potential for better performance, these methods contradict our purpose.
Therefore, they are not included in the fundamental comparison for fairness.
To illustrate the value of studying parameter-efficient methods, in \cref{tab:results_freeze_CLIP}, we compare with the state-of-the-art X-Pool~\cite{Gorti:X-Pool} by freezing the CLIP backbone.
We observe that without fine-tuning the backbone, X-Pool underperforms our VoP$^\text{F}$ with much more parameter overhead.
And equipping our simplest VoP significantly boosts its performance with negligible additional parameters.
The comparison results demonstrate the superiority of our proposed methods as parameter-efficient solutions.

\section{Retrieval Results with ViT-B/16}

\begin{table}[th]
\tablestyle{5pt}{1.0}
\setlength\tabcolsep{2.5pt}
\def\w{20pt} 
\scalebox{1}{
  \begin{tabular}{lc|cccp{0.7cm}<{\centering}p{0.85cm}<{\centering}}
    \textbf{Methods} & \textbf{Params (M)} & \textbf{R@1} & \textbf{R@5} & \textbf{R@10} & \textbf{MnR↓} & \textbf{MdR↓} \\
    \shline
    Full & 118.1 (100\%) & 44.9  & 72.2  & 81.7  & 13.5  & 2.0   \\
    Bias~\cite{Cai:ft-bias} & 0.1 (0.105\%) & 42.2  & 68.5  & 78.2  & 13.9  & 2.0   \\
    Proj~\cite{Jia:VPT} & 0.7 (0.555\%) & 39.1  & 65.7  & 75.6  & 17.6  & 2.0   \\
    Partial~\cite{Jia:VPT} & 7.7 (6.506\%) & 43.0  & 69.3  & 78.5  & 15.8  & 2.0   \\
    Adapter$^\text{ATTN}$~\cite{He:unify-PET} & 2.0 (1.680\%) & 41.7  & 66.4  & 76.6  & 15.1  & 2.0   \\
    Adapter$^\text{FFN}$~\cite{Chen:AdaptFormer} & 2.0 (1.680\%) & 41.4  & 66.5  & 77.0  & 15.0  & 2.0   \\
    Ju \textit{et al.}~\cite{Ju:prompt-for-video} & 4.8 (3.990\%) & 36.7 & 64.6 & 76.8 & - & 2.0 \\
    \hline
    \textbf{VoP} & 0.1 (0.104\%) & 43.4  & 69.1  & 80.5  & 14.2  & 2.0   \\
    \textbf{VoP$^\text{P}$} & 0.5 (0.448\%) & 43.9  & 70.0  & 80.9  & 12.9  & 2.0   \\
    \textbf{VoP$^\text{C}$} & 14.3 (12.077\%) & 44.6  & 71.8  & 80.2  & 14.6  & 2.0   \\
    \textbf{VoP$^\text{F}$} & 0.1 (0.104\%) & 46.5  & \textbf{73.0}  & 81.5  & \uline{12.4}  & 2.0   \\
    \hline
    \specialrule{0em}{0.5pt}{0.5pt}
    \textbf{VoP$^\text{F+P}$} & 0.4 (0.333\%) & \uline{47.1}  & \uline{72.4} & \uline{81.8}  & 12.9  & 2.0   \\
    \textbf{VoP$^\text{F+C}$} & 14.1 (11.962\%) & \textbf{47.7} & \uline{72.4} & \textbf{82.2} & \textbf{12.0} & 2.0   \\
  \end{tabular}
  }
  \vspace{-2mm}
  \caption{\textbf{$t2v$ results on the MSR-VTT-9k dataset with ViT-B/16}.}
  \label{tab:results_ViT-B-16}%
  \vspace{-2mm}
\end{table}

In this section, we change the visual encoder to a ViT-B/16 to examine %
all solutions including ours with a heavier backbone.
Compared to the default ViT-B/32, ViT-B/16 splits the image into more and smaller 16$\times$16 patches, increasing the computational effort to learn more detailed relational information while slightly reducing the number of parameters (118.1M \textit{v.s.}119.8M).
We here report the $t2v$ results obtained on MSR-VTT-9k in \cref{tab:results_ViT-B-16} and also compare with the method proposed by Ju \textit{et al.}~\cite{Ju:prompt-for-video}.
Several observations as follows:
(1) Our VoP now outperforms all parameter-efficient tuning protocols including Partial, showing its ability to effectively transfer the latent knowledge with fewer trainable parameters.
(2) The proposed video prompts still steadily reinforce VoP, where VoP$^\text{F}$ and its variants outperform Full.
(3) equipping with two video prompts brings a 3.7\% to 4.3\% improvement to VoP, and our VoP$^\text{F+C}$ even yields a remarkable $t2v$ R@1 47.7\%.

\section{Detailed Retrieval Results}

We here report the detailed retrieval results on MSR-VTT-7k (\cref{tab:main_results_MSRVTT7k}), DiDeMo (\cref{tab:main_results_DiDeMo}), ActivityNet (\cref{tab:main_results_ActivityNet}), LSMDC (\cref{tab:main_results_LSMDC}) for reference.
Note that these results are obtained using CLIP with ViT-B/32 unless otherwise stated.
The conclusions in these tables are generally consistent with those from the above experiments.

\begin{table*}[!t]    %
\tablestyle{5pt}{1.0}
\setlength\tabcolsep{4pt}
\def\w{20pt} 
\scalebox{1}{
    \begin{tabular}{lc|cccp{0.7cm}<{\centering}p{0.85cm}<{\centering}|cccp{0.7cm}<{\centering}p{0.75cm}<{\centering}}
    \multirow{2}[1]{*}{\textbf{Methods}} & \multirow{2}[1]{*}{\textbf{Params (M)}} & \multicolumn{5}{c}{\textbf{$t2v$}}     & \multicolumn{5}{c}{\textbf{$v2t$}} \\
          &       & \textbf{R@1} & \textbf{R@5} & \textbf{R@10} & \textbf{MnR↓} & \textbf{MdR↓} & \textbf{R@1} & \textbf{R@5} & \textbf{R@10} & \textbf{MnR↓} & \textbf{MdR↓} \\
    \shline
    Full  & 119.8 (100\%) & 40.9  & 67.9  & 78.4  & 18.3  & 2.0 & 41.7  & \uline{69.6}  & 79.7  & 12.7  & 2.0 \\
    Bias~\cite{Cai:ft-bias}  & 0.1 (0.104\%) & 39.7  & 65.9  & 76.7  & 17.9  & 2.0 & 41.2  & 66.6  & 78.9  & 14.0  & 2.0 \\
    Proj~\cite{Jia:VPT}  & 0.7 (0.547\%) & 36.0  & 63.6  & 74.6  & 21.4  & 3.0 & 36.9  & 63.6  & 74.6  & 17.8  & 3.0 \\
    Partial~\cite{Jia:VPT} & 7.7 (6.410\%) & 39.2  & 64.0  & 74.7  & 20.9  & 3.0 & 37.7  & 63.6  & 74.9  & 16.9  & 3.0 \\
    Adapter$^\text{ATTN}$~\cite{He:unify-PET} & 2.0 (1.655\%) & 39.6  & 65.4  & 76.8  & 16.8  & 2.0 & 41.6  & 67.6  & 79.8  & 12.4  & 2.0 \\
    Adapter$^\text{FFN}$~\cite{Chen:AdaptFormer} & 2.0 (1.655\%) & 39.9  & 65.3  & 76.9  & 16.8  & 2.0 & 41.6  & 67.6  & 79.2  & 12.7  & 2.0 \\
    \hline
    \textbf{VoP} & 0.1 (0.103\%) & 39.7  & 66.7  & 77.9  & 16.7  & 2.0 & 41.4  & 68.8  & \textbf{80.8}  & 12.5  & 2.0 \\
    \textbf{VoP$^\text{P}$} & 0.5 (0.441\%) & 40.6  & 66.0  & 76.7  & 16.6  & 2.0 & 41.6  & 69.0  & 79.5  & 12.3  & 2.0 \\
    \textbf{VoP$^\text{C}$} & 14.3 (11.898\%) & 40.0  & 67.3  & 78.2  & 17.0  & 2.0 & 41.7  & 69.4  & 79.1  & 12.3  & 2.0 \\
    \textbf{VoP$^\text{F}$} & 0.1 (0.103\%) & 42.0  & 67.4  & 78.2  & 16.2  & 2.0 & 42.8  & 68.4  & 79.8  & 12.3  & 2.0 \\
    \hline
    \specialrule{0em}{0.5pt}{0.5pt}
    \textbf{VoP$^\text{F+P}$} & 0.4 (0.328\%) & \textbf{43.5} & \uline{68.1}  & \uline{79.2}  & \uline{16.0}  & 2.0 & \uline{43.4}  & \textbf{71.0}  & \uline{80.4}  & \textbf{11.3}  & 2.0 \\
    \textbf{VoP$^\text{F+C}$} & 14.1 (11.785\%) & \uline{42.7}  & \textbf{68.2} & \textbf{79.3} & \textbf{15.9} & 2.0 & \textbf{44.2}  & \uline{69.6}  & \textbf{80.8}  & \uline{11.4}  & 2.0 \\
    \end{tabular}%
    }
    \vspace{-2mm}
  \caption{\textbf{Retrieval results on the MSR-VTT-7k dataset}. 
  }
  \label{tab:main_results_MSRVTT7k}%
  \vspace{-3mm}
\end{table*}%

\begin{table*}[!t]    %
\tablestyle{5pt}{1.0}
\setlength\tabcolsep{4pt}
\def\w{20pt} 
\scalebox{1}{
    \begin{tabular}{lc|cccp{0.7cm}<{\centering}p{0.85cm}<{\centering}|cccp{0.7cm}<{\centering}p{0.75cm}<{\centering}}
    \multirow{2}[1]{*}{\textbf{Methods}} & \multirow{2}[1]{*}{\textbf{Params (M)}} & \multicolumn{5}{c}{\textbf{$t2v$}}     & \multicolumn{5}{c}{\textbf{$v2t$}} \\
          &       & \textbf{R@1} & \textbf{R@5} & \textbf{R@10} & \textbf{MnR↓} & \textbf{MdR↓} & \textbf{R@1} & \textbf{R@5} & \textbf{R@10} & \textbf{MnR↓} & \textbf{MdR↓} \\
    \shline
    Full  & 119.8 (100\%) & 41.6  & 68.4  & 78.2  & 17.7  & 2.0 & 40.2  & 68.4  & 78.7  & 11.9  & 2.0  \\
    Bias~\cite{Cai:ft-bias}  & 0.1 (0.104\%) & 36.5  & 63.4  & 75.2  & 24.8  & 3.0 & 36.8  & 65.7  & 75.8  & 15.1  & 2.0  \\
    Proj~\cite{Jia:VPT}  & 0.7 (0.547\%) & 35.6  & 61.3  & 72.6  & 24.4  & 3.0 & 34.5  & 60.9  & 72.6  & 18.8  & 3.0  \\
    Partial~\cite{Jia:VPT} & 7.7 (6.410\%) & 39.3  & 65.5  & 75.7  & 22.3  & 2.0 & 36.9  & 64.2  & 74.5  & 17.0  & 2.0   \\
    Adapter$^\text{ATTN}$~\cite{He:unify-PET} & 2.0 (1.655\%) & 36.4  & 62.8  & 73.9  & 23.5  & 3.0 & 36.3  & 64.4  & 74.8  & 15.4  & 2.0   \\
    Adapter$^\text{FFN}$~\cite{Chen:AdaptFormer} & 2.0 (1.655\%) & 36.3  & 63.4  & 75.4  & 22.9  & 3.0 & 35.6  & 64.3  & 75.6  & 14.8  & 3.0   \\
    \hline
    \textbf{VoP} & 0.1 (0.103\%) & 38.2  & 66.9  & 76.1  & 19.8  & 2.0 & 38.1  & 65.7  & 76.5  & 13.5  & 2.0  \\
    \textbf{VoP$^\text{P}$} & 0.5 (0.441\%) & 38.9  & 67.7  & 78.1  & 17.2  & 2.0 & 40.6  & 68.3  & 78.6  & 11.6  & 2.0  \\
    \textbf{VoP$^\text{C}$} & 14.3 (11.898\%) & 40.0  & 68.0  & 78.5  & 18.3  & 2.0 & 39.1  & 65.3  & 76.7  & 13.8  & 3.0  \\
    \textbf{VoP$^\text{F}$} & 0.1 (0.103\%) & 44.7  & 70.8  & 79.7  & 15.7  & 2.0 & 43.5  & 70.9  & \uline{81.4}  & \uline{9.8}  & 2.0  \\
    \hline
    \specialrule{0em}{0.5pt}{0.5pt}
    \textbf{VoP$^\text{F+P}$} & 0.4 (0.328\%) & \uline{45.3}  & \textbf{72.3} & \uline{80.4}  & \uline{13.8}  & 2.0 & \textbf{44.7}  & \uline{71.2}  & 81.1  & 9.9  & 2.0  \\
    \textbf{VoP$^\text{F+C}$} & 14.1 (11.785\%) & \textbf{46.4} & \uline{71.9}  & \textbf{81.5} & \textbf{13.6} & 2.0 & \uline{44.4}  & \textbf{71.8}  & \textbf{81.8}  & \textbf{9.5}  & 2.0  \\
    \end{tabular}%
    }
    \vspace{-2mm}
  \caption{\textbf{Retrieval results on the DiDeMo dataset}. 
  }
  \label{tab:main_results_DiDeMo}%
  \vspace{-3mm}
\end{table*}%

\begin{table*}[!t]    %
\tablestyle{5pt}{1.0}
\setlength\tabcolsep{4pt}
\def\w{20pt} 
\scalebox{1}{
    \begin{tabular}{lc|cccp{0.7cm}<{\centering}p{0.85cm}<{\centering}|cccp{0.7cm}<{\centering}p{0.75cm}<{\centering}}
    \multirow{2}[1]{*}{\textbf{Methods}} & \multirow{2}[1]{*}{\textbf{Params (M)}} & \multicolumn{5}{c}{\textbf{$t2v$}}     & \multicolumn{5}{c}{\textbf{$v2t$}} \\
          &       & \textbf{R@1} & \textbf{R@5} & \textbf{R@10} & \textbf{MnR↓} & \textbf{MdR↓} & \textbf{R@1} & \textbf{R@5} & \textbf{R@10} & \textbf{MnR↓} & \textbf{MdR↓} \\
    \shline
    Full  & 119.8 (100\%) & \textbf{36.8} & \textbf{66.9} & \textbf{80.1} & \textbf{9.3} & 3.0 & \textbf{38.9}  & \textbf{70.1}  & \textbf{81.9}  & \textbf{8.4}  & 2.0   \\
    Bias~\cite{Cai:ft-bias}  & 0.1 (0.104\%) & 31.3  & 60.3  & 74.2  & 13.4  & 3.0 & 33.7  & 63.8  & 77.6  & 11.4  & 3.0  \\
    Proj~\cite{Jia:VPT}  & 0.7 (0.547\%) & 29.8  & 59.1  & 73.3  & 14.2  & 4.0 & 31.1  & 60.6  & 74.6  & 13.1  & 3.0   \\
    Partial~\cite{Jia:VPT} & 7.7 (6.410\%) & 33.6  & 64.0  & 77.8  & \uline{10.6}  & 3.0 & 33.4  & 64.6  & 77.8  & 10.2  & 3.0   \\
    Adapter$^\text{ATTN}$~\cite{He:unify-PET} & 2.0 (1.655\%) & 31.6  & 60.5  & 74.4  & 13.1  & 3.0 & 33.3  & 63.6  & 77.1  & 11.3  & 3.0   \\
    Adapter$^\text{FFN}$~\cite{Chen:AdaptFormer} & 2.0 (1.655\%) & 31.8  & 61.0  & 75.0  & 12.8  & 3.0 & 33.6  & 63.9  & 77.3  & 11.1  & 3.0   \\
    \hline
    \textbf{VoP} & 0.1 (0.103\%) & 32.3  & 61.9  & 75.5  & 12.4  & 3.0 & 33.7  & 64.7  & 77.2  & 11.1  & 3.0   \\
    \textbf{VoP$^\text{P}$} & 0.5 (0.441\%) & 32.8  & 62.3  & 75.4  & 12.3  & 3.0 & 34.8  & 65.0  & 78.2  & 10.7  & 3.0   \\
    \textbf{VoP$^\text{C}$} & 14.3 (11.898\%) & 32.6  & 62.5  & 76.5  & 12.0  & 3.0 & 34.2  & 64.8  & 78.4  & 10.7  & 3.0   \\
    \textbf{VoP$^\text{F}$} & 0.1 (0.103\%) & 34.6  & 62.6  & 76.4  & 11.6  & 3.0 & 35.5  & 65.1  & 77.4  & 10.2  & 3.0   \\
    \hline
    \specialrule{0em}{0.5pt}{0.5pt}
    \textbf{VoP$^\text{F+P}$} & 0.4 (0.328\%) & \uline{36.1}  & \uline{65.5}  & \uline{78.5}  & 10.9  & 3.0 & \uline{36.3}  & \uline{65.9}  & \uline{79.2}  & \uline{10.1}  & 3.0  \\
    \textbf{VoP$^\text{F+C}$} & 14.1 (11.785\%) & 35.1  & 63.7  & 77.6  & 11.4  & 3.0 & 35.6  & \uline{65.9}  & 77.8  & 10.4  & 3.0   \\
    \end{tabular}%
    }
    \vspace{-2mm}
  \caption{\textbf{Retrieval results on the ActivityNet dataset}. 
  }
  \label{tab:main_results_ActivityNet}%
  \vspace{-3mm}
\end{table*}%

\begin{table*}[!t]    %
\tablestyle{5pt}{1.0}
\setlength\tabcolsep{4pt}
\def\w{20pt} 
\scalebox{1}{
    \begin{tabular}{lc|cccp{0.7cm}<{\centering}p{0.85cm}<{\centering}|cccp{0.7cm}<{\centering}p{0.75cm}<{\centering}}
    \multirow{2}[1]{*}{\textbf{Methods}} & \multirow{2}[1]{*}{\textbf{Params (M)}} & \multicolumn{5}{c}{\textbf{$t2v$}}     & \multicolumn{5}{c}{\textbf{$v2t$}} \\
          &       & \textbf{R@1} & \textbf{R@5} & \textbf{R@10} & \textbf{MnR↓} & \textbf{MdR↓} & \textbf{R@1} & \textbf{R@5} & \textbf{R@10} & \textbf{MnR↓} & \textbf{MdR↓} \\
    \shline
    Full  & 119.8 (100\%) & \textbf{22.0} & 39.9  & \textbf{49.9} & \textbf{56.8} & 11.0 & \uline{21.9}  & 40.0  & 48.2  & \textbf{50.7}  & 12.0   \\
    Bias~\cite{Cai:ft-bias}  & 0.1 (0.104\%) & 17.4  & 36.2  & 44.9  & 73.2  & 14.0 & 18.0  & 36.0  & 44.9  & 62.2  & 15.0   \\
    Proj~\cite{Jia:VPT}  & 0.7 (0.547\%) & 15.7  & 32.7  & 40.8  & 83.7  & 20.0 & 17.1  & 32.6  & 39.9  & 76.4  & 21.0   \\
    Partial~\cite{Jia:VPT} & 7.7 (6.410\%) & 18.0  & 33.8  & 41.8  & 79.9  & 18.0 & 15.9  & 33.2  & 41.5  & 72.3  & 18.0   \\
    Adapter$^\text{ATTN}$~\cite{He:unify-PET} & 2.0 (1.655\%) & 18.4  & 38.0  & 46.4  & 68.9  & 13.0 & 19.7  & 37.6  & 46.3  & 55.4  & 13.0  \\
    Adapter$^\text{FFN}$~\cite{Chen:AdaptFormer} & 2.0 (1.655\%) & 18.7  & 38.9  & 47.3  & 63.6  & 13.0 & 19.8  & 38.4  & 47.0  & 57.8  & 12.0   \\
    Ju \textit{et al.}~\cite{Ju:prompt-for-video} $^\dag$ & 4.8 (3.990\%) & 18.8  & 38.5  & 47.9  & - & 12.3 & - & - & - & - & - \\
    \hline
    \textbf{VoP} & 0.1 (0.103\%) & 19.0  & 37.9  & 46.5  & 66.9  & 14.0 & 18.5  & 36.1  & 45.3  & 59.5  & 14.0  \\
    \textbf{VoP$^\text{P}$} & 0.5 (0.441\%) & 19.2  & 38.3  & 47.3  & 64.4  & 12.0 & 19.7  & 38.9  & 48.1  & 55.4  & 12.0   \\
    \textbf{VoP$^\text{C}$} & 14.3 (11.898\%) & 20.4  & 40.0  & 48.1  & 65.9  & 12.0 & 20.3  & 38.7  & 48.5  & 56.9  & 11.0   \\
    \textbf{VoP$^\text{F}$} & 0.1 (0.103\%) & 20.6  & 39.5  & 49.1  & 60.3  & 11.0 & 21.2  & 39.4  & \uline{49.2}  & 52.3  & 11.0  \\
    \hline
    \specialrule{0em}{0.5pt}{0.5pt}
    \textbf{VoP$^\text{F+P}$} & 0.4 (0.328\%) & 20.7  & \uline{40.7}  & \uline{49.7}  & \uline{59.1}  & 11.0 & 21.5  & \textbf{40.6}  & \textbf{50.7}  & \uline{50.8}  & 10.0  \\
    \textbf{VoP$^\text{F+C}$} & 14.1 (11.785\%) & \uline{21.1}  & \textbf{40.9} & 49.6  & 60.1  & 11.0 & \textbf{22.3}  & \uline{40.3}  & \textbf{50.7}  & 51.1  & 10.0 \\
    \end{tabular}%
    }
    \vspace{-2mm}
  \caption{\textbf{Retrieval results on the LSMDC dataset}. $^\dag$~denotes that it uses CLIP with ViT-B/16.
  }
  \label{tab:main_results_LSMDC}%
  \vspace{-3mm}
\end{table*}%

\end{document}